\documentclass[11pt,a4paper]{article}
\usepackage[hyperref]{emnlp-ijcnlp-2019}
\usepackage{times}
\usepackage{latexsym}

\usepackage{algorithm}
\usepackage{algorithmic}
\usepackage{url}
\usepackage{graphicx}
\usepackage{caption}
\usepackage{subcaption}
\usepackage{textgreek}
\usepackage{booktabs}
\usepackage{multirow}
\usepackage{todonotes}
\usepackage{array}
\usepackage{textcomp}
\usepackage{multirow}
\usepackage{booktabs}
\usepackage{graphicx}
\usepackage{array}
\usepackage{marvosym}
\usepackage{rotating}

\graphicspath{ {./images/} }

\newcolumntype{s}{>{\small}c}
\newcolumntype{k}{>{\small}l}
\newcolumntype{j}{>{\small}r}

\aclfinalcopy 

\usepackage{amsmath}
\usepackage{amssymb}
\newcommand{\mathbold}[1]{\ensuremath{\boldsymbol{\mathbf{#1}}}}

\DeclareRobustCommand{\mb}[1]{\ensuremath{\boldsymbol{\mathbf{#1}}}}

\newcommand{\mbw}{\mathbold{w}}
\newcommand{\mbx}{\mathbold{x}}
\newcommand{\mby}{\mathbold{y}}

\newcommand{\mbH}{\mathbold{H}}
\newcommand{\mbI}{\mathbold{I}}

\newcommand{\mbK}{\mathbold{K}}
\newcommand{\mbL}{\mathbold{L}}

\newcommand{\mbW}{\mathbold{W}}
\newcommand{\mbX}{\mathbold{X}}
\newcommand{\mbY}{\mathbold{Y}}

\DeclareMathOperator*{\Tr}{Tr}

\newcommand{\mbbR}{\mathbb{R}}

\newcommand{\RomanNumeralCaps}[1]
    {\MakeUppercase{\romannumeral #1}}

\usepackage{cleveref}  

\title{Correlations between Word Vector Sets}

\author{Vitalii Zhelezniak, April Shen, Daniel Busbridge, \\
\smallskip
\textbf{Aleksandar Savkov,} \textbf{Nils Hammerla}
 \\
\texttt{\{firstname.lastname\}}
\texttt{@babylonhealth.com}
}

\date{}

\begin{document}
\maketitle

\begin{abstract}

Similarity measures based purely on word embeddings are comfortably competing with much more
sophisticated deep learning and expert-engineered systems on unsupervised semantic textual similarity (STS) tasks.
In contrast to commonly used geometric approaches, we treat a single word embedding as e.g.\ 300
observations from a scalar random variable.
Using this paradigm, we first illustrate that similarities derived from elementary pooling operations and classic
correlation coefficients yield excellent results on standard STS benchmarks, outperforming many recently proposed
methods while being much faster and trivial to implement.
Next, we demonstrate how to avoid pooling operations altogether and compare sets of word embeddings directly via
correlation operators between reproducing kernel Hilbert spaces.
Just like cosine similarity is used to compare individual word vectors, we introduce a novel application of the centered
kernel alignment (CKA) as a natural generalisation of squared cosine similarity for sets of word vectors.
Likewise, CKA is very easy to implement and enjoys very strong empirical results.

\end{abstract}

\section{Introduction}
\label{sec:intro}

Distributed representations of text have had a massive impact on the natural language
processing (NLP), information retrieval (IR), and machine learning (ML) communities, thanks in part to their ability to capture rich notions
of semantic similarity.
While this work originally began with word embeddings \citep{Bengio2003, Mikolov2013, Pennington2014, Bojanowski2016, Joulin2016a},
there is now an ever-increasing number of representations for longer units of text based on simple aggregations of word vectors \citep{Mitchell2008, DeBoom2016, Arora2017, Wieting2016, Wieting2018, zhelezniak2019a}
as well as complex neural architectures \citep{Le2014, Kiros2015, Hill2016, Conneau2017, Gan2017, Tang2017, Pagliardini2018, Zhelezniak2018, Subramanian2018, USE2018, devlin2018bert}.

By contrast, relatively little effort has been directed towards understanding the similarity measures used to compare these textual embeddings,
for which cosine similarity remains a convenient and widespread, yet somewhat arbitrary default, despite some emerging research into the alternatives
\citep{Camacho-Collados2015, DeBoom2015, Santus2018, zhelezniak2019a, zhelezniak2019b}.
Part of the appeal of cosine similarity perhaps lies in the simple geometric interpretation behind it.
However, as embeddings are ultimately just arrays of numbers, we are free to take alternative viewpoints other than the geometric ones, if they lead to illuminating insights or strong-performing methods.

Following \citet{zhelezniak2019b}, we treat a word embedding not as a geometric vector but as a statistical sample (of e.g.\ 300
observations) from a scalar random variable, and indeed find insights that are both intriguing and noteworthy.
We first illustrate that similarities derived from elementary pooling operations and
classic univariate correlation coefficients yield excellent results on standard semantic textual similarity (STS) benchmarks, outperforming
many recently proposed methods while being much faster and simpler to implement.
This empirically validates the advantages of the statistical perspective on word embeddings over the geometric interpretations.
In the process, we provide more evidence that departures from normality, and in particular the presence of outliers, can have severe negative effects on the performance of some correlation coefficients.
We show how to overcome these complications, by selecting an outlier-removing pooling operation such as max-pooling, applying a more robust correlation coefficient such as Spearman's $\rho$, or simply clipping (winsorizing) the word vectors.

Next, we demonstrate how to avoid pooling operations completely and compare sets of word embeddings directly via
correlation operators between reproducing kernel Hilbert spaces (RKHS).
We introduce a novel application of the kernel alignment (KA) and the centered kernel alignment (CKA) as a natural generalisation of the squared cosine similarity and Pearson correlation
for the sets of word embeddings.
These multivariate correlation coefficients are very easy to implement and also enjoy very strong empirical results.

\section{Related Work}
\label{sec:related}

Several lines of research seek to combine the strength of pretrained word embeddings and the elegance of
set- or bag-of-words (BoW) representations.
Any method that determines semantic similarity between sentences by comparing the corresponding sets of word embeddings is directly related to our work.

Perhaps the most obvious such approaches are based on elementary pooling operations such as average-, max- and min-pooling~\citep{Mitchell2008, DeBoom2015, DeBoom2016}.
While seemingly over-simplistic, numerous studies have confirmed their impressive performance on the downstream tasks~\citep{Arora2017, Wieting2016, Wieting2018, zhelezniak2019a}

One step further, \citet{Zhao2017, zhelezniak2019a} introduce fuzzy bags-of-words (FBoW) where degrees of membership in a fuzzy set are given by the similarities between word embeddings.
\citet{zhelezniak2019a} show a close connection between FBoW and max-pooled word vectors.

Some approaches do not seek to build an explicit representation and instead focus directly on designing a similarity function between sets.
Word Mover's Distance (WMD) \citep{Kusner2015} is an instance of the Earth Mover's Distance (EMD) computed between normalised BoW, with the cost matrix given by Euclidean distances between word embeddings.
In the soft cardinality framework of \citep{Jimenez2010, Jimenez2015}, the contribution of a word to the cardinality of a set depends on its similarities to other words in the same set.
Such sets are then compared using an appropriately defined Jaccard index or related measures.
DynaMax \citep{zhelezniak2019a} uses universe-constrained fuzzy sets designed explicitly for similarity computations.

Approaches that see word embeddings as statistical objects are very closely related to our work.
Virtually all of them treat word embeddings as observations from some $D$-variate parametric family, where $D$ is the embedding dimension.
\citet{Arora2016,Arora2017} introduce a latent discourse model and show the maximum likelihood estimate (MLE) for the discourse vector to be the weighted average of word embeddings in a sentence, where the weights are given by smooth inverse frequencies (SIF).
\citet{nikolentzos2017, torki2018} treat sets of word embeddings as observations from $D$-variate Gaussians, and compare such sets with cosine similarity between the parameters (means and covariances) estimated by maximum likelihood.
\citet{vargas2019} measure semantic similarity through penalised likelihood ratio between the joint and factorised models and explore Gaussian and von Mises--Fisher likelihoods.

Cosine similarity between covariances is an instance of the RV coefficient and its uncentered version was applied in the context of word embeddings before \citep{botev2017}.
We arrive at a similar coefficient (but with different centering) as a special case of CKA, which in the general case makes no parametric assumptions about disbtributions whatsoever.
In particular our version is suitable for comparing sets containing just one word vector, whereas the method of \citet{nikolentzos2017, torki2018} requires at least two vectors in each set.
Very recently, \citet{kornblith2019} used CKA to compare representations between layers of the same or different neural networks.
This is again an instance of treating such representations as observations from a $D$-variate distribution, where $D$ is the dimension of the hidden layer in question.
Our use of CKA is completely different from theirs.

Unlike all of the above approaches, \citep{zhelezniak2019b} see each word embedding itself as $D$ (e.g.\ 300) observations from some scalar random variable.
They cast semantic similarity as correlations between these random variables and study their properties using simple tools from univariate statistics.
While they consider correlations between individual word vectors and averaged word vectors, they do not formally explore correlations between word vector sets.
We review their framework in \Cref{sec:corrsim} and then proceed to formalise and generalise it to the case of sets of word embeddings.

\section{Background: Correlation Coefficients and Semantic Similarity}
\label{sec:corrsim}

Suppose we have a word embeddings matrix $\mbW \in \mathbb{R}^{N \times D}$, where $N$ is the number of words in the vocabulary and $D$ is the embedding dimension (usually 300).
In other words, each row $\mbw^{(i)}$ of $\mbW$ is a $D$-dimensional word vector.
When applying statistical analysis to these vectors, one might choose to treat each $\mbw^{(i)}$ as an observation from some $D$-variate distribution $P_D(E_1, \ldots E_{D})$ and model it with a Gaussian or a Gaussian Mixture.
While such analysis helps in studying the overall geometry of the embedding space (how dimensions correlate and how embeddings cluster), $P_D$ is not directly useful for semantic similarity between individual words.

For the latter, \citet{zhelezniak2019b} proposed to look at the transpose $\mbW^{T}$ and the corresponding distribution $P(W_1, W_2, \ldots, W_N)$.
Under this perspective, each word vector $\mbw_{(i)}$ is now a sample of $D$ (e.g.\ 300) observations from a scalar random variable $W_i$.
Luckily, in applications we are usually not interested in the full joint distribution but only in the similarity between two words, i.e.\ the bivariate marginal $P(W_i, W_j)$.
In practice, we make inferences about this marginal from the paired sample $(\mbw_{(i)}, \mbw_{(j)})$ through visualisations (histograms, Q-Q plots, scatter plots, etc.) as well as various statistics.

\citet{zhelezniak2019b} found that for all common models (GloVe, fastText, word2vec) the means across word embeddings are tightly concentrated around zero (relative to their dimensions), thus making the widely used cosine similarity practically equivalent to Pearson correlation.
However, while word2vec vectors seem mostly normal, GloVe and fastText vectors are highly non-normal, likely due to the presence of heavy univariate and bivariate outliers (as suggested by visualisations mentioned earlier).
Quantitatively, the majority of GloVe and fastText vectors fail the Shapiro-Wilk normality test at significance level 0.05.
Therefore, while Pearson's $r$ (and thus cosine similarity) may be acceptable for word2vec, it is preferable to resort to more robust non-parametic correlation coefficients such as Spearman's $\rho$ or Kendall's $\tau$ as a similarity measure between GloVe and fastText vectors.

Finally, very similar conclusions were shown to hold for sentence representations obtained by word vector averaging, also referred to as mean-pooling.
In particular, averaged fastText vectors compared with rank correlation coefficients already show impressive results on standard STS tasks, rivaling much more sophisticated systems.

\section{Correlations between Word Vector Sets}
\label{sec:theory}

We are interested in applying the statistical framework from \Cref{sec:corrsim} to measure the semantic similarity between two sentences $s_1$ and $s_2$ given by the sets (or bags) $S_1$ and $S_2$ of word embeddings respectively.
To formalise this new setup, we may see each set of word embeddings $S = \{\mbw_{(1)}, \mbw_{(2)}, \ldots, \mbw_{(k)}\}$ as a sample (of e.g. 300 observations) from some theoretical set of scalar random variables $R = \{W_1, W_2, \ldots, W_k\}$.
In light of the above, our task then lies in finding correlation coefficients $\text{corr}(R_1, R_2)$ between $R_1$ and $R_2$ and their empirical estimates $\widehat{\text{corr}}(S_1, S_2)$ obtained from the paired sample $S_1, S_2$, hoping that such coefficients will serve as a good proxy for semantic similarity.
Recall that for single-word sets $R_1 = \{W_i\}, R_2 = \{W_j\}$ the task simply reduces to computing a univariate correlation between word vectors $\mbw_{(i)}$ and $\mbw_{(j)}$, where the choice of the coefficient (Pearson's $r$, Spearman's $\rho$, etc.) is made based on the statistics exhibited by the word embeddings matrix.
While generalising this to sets of more than one variable is not particularly hard, there are several ways to do so, each with its own advantages and downsides.
In the present work, we group these approaches into two broad families: pooling-based and pooling-free correlation coefficients.

\begin{figure*}[h]
	\centering
	\includegraphics[width=\textwidth]{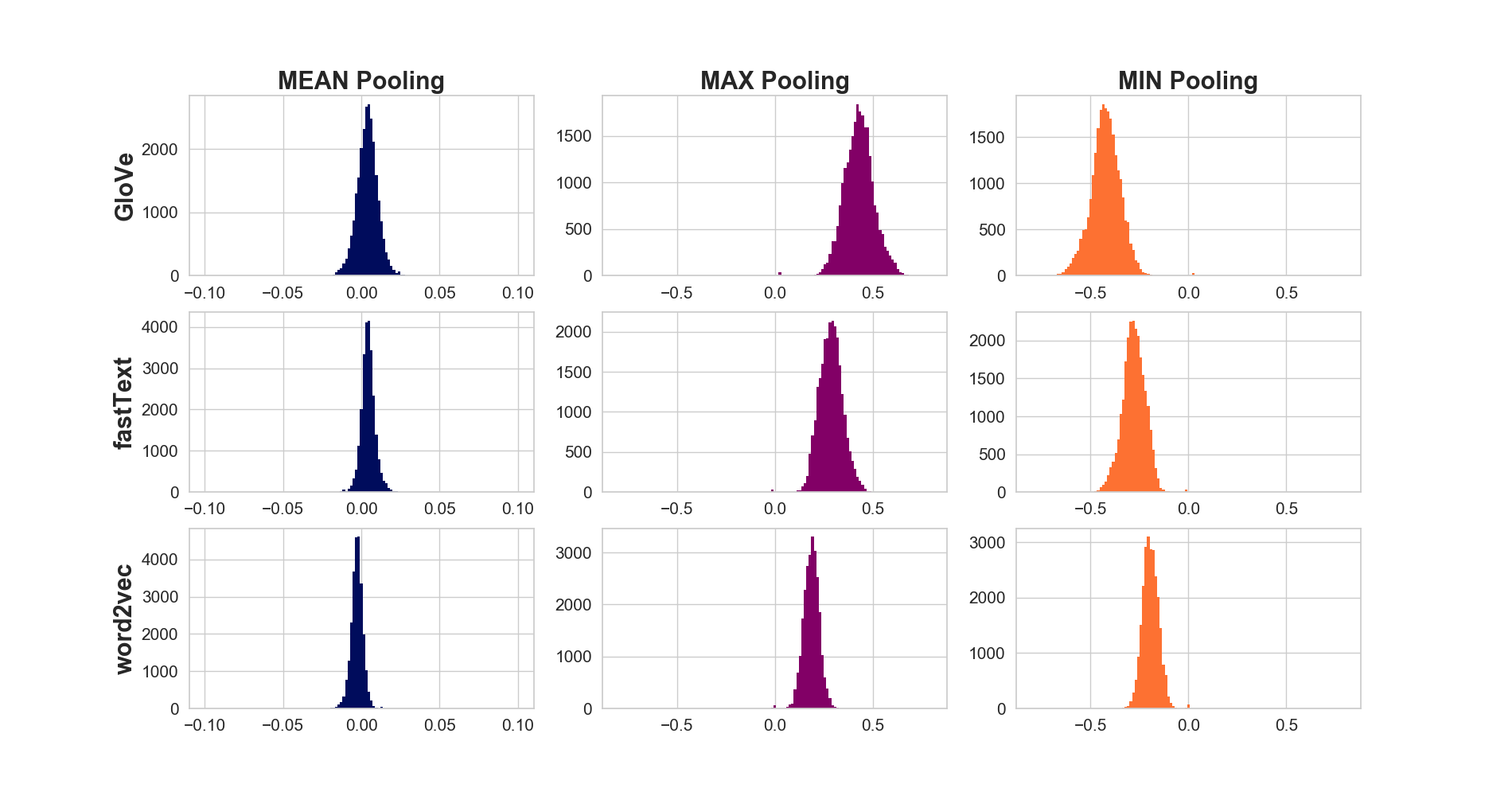}
	\caption{
	Normalised histograms of the mean distribution for sentence vectors generated by mean-, max- and min-pooling.
	Sentences were taken from the entire STS dataset \citep{Agirre2012, Agirre2013a, Agirre2014, Agirre2015, Agirre2016, Cer2017},
	and we utilise three commonly-used word embedding models:
	GloVe~\cite{Pennington2014}, fastText~\cite{Bojanowski2016}, and word2vec~\cite{Mikolov2013a,Mikolov2013b}.
	}
	\label{fig:means}
\end{figure*}

\begin{algorithm}[t]
\caption{MaxPool-Spearman}
\label{alg:maxpool}
\begin{algorithmic}
\REQUIRE Word embeddings for the first sentence $\mbx^{(1)}, \mbx^{(2)} \ldots, \mbx^{(k)} \in \mbbR^{1\times d}$
\REQUIRE Word embeddings for the second sentence $\mby^{(1)}, \mby^{(2)} \ldots, \mby^{(l)} \in \mbbR^{1\times d}$
\ENSURE Similarity score $MS$

\STATE \# \textit{Max-pooling performed element-wise}
\STATE $\mbx \leftarrow \textsc{max\_pool}( \mbx^{(1)}, \mbx^{(2)} \ldots, \mbx^{(k)} )$
\STATE $\mby \leftarrow \textsc{max\_pool}( \mby^{(1)}, \mby^{(2)} \ldots, \mby^{(l)} )$

\STATE $\textsc{MS} \leftarrow \textsc{SpearmanCorrelation}(\mbx, \mby)$
\end{algorithmic}

\end{algorithm}

\subsection{Correlations between Pooled Variables}
\label{subsec:correlationsBetweenPooledVariables}

Pooling-based approaches first reduce a set of random variables to a single scalar random variable $W_{\text{pool}} = f_\text{{pool}}(W_1, W_2, \ldots, W_k)$ and then apply univariate correlation coefficients between the pooled variables.
In practice this would correspond to pooling word embeddings $\mbw_{(1)}, \mbw_{(2)}, \ldots, \mbw_{(k)}$ (along $i=1{:}k$) into one fixed vector $\mbw_{\text{pool}}$, followed by computing univariate sample correlations.
Certainly, these approaches are empirically attractive: not only are they very simple computationally (e.g.\ see~\Cref{alg:maxpool}) but they also keep us in the realm of univariate statisics, where we have an entire arsenal of effective tools for making inferences about $W_{\text{pool}}$.

Unfortunately, it is not always clear \emph{a priori} what should dictate our choice of the pooling function (though, as we will see shortly, for certain functions some statistical justifications do exist).
By far the most common pooling operations for word embedding found in the literature are \mbox{mean-,} max- and min-pooling.
It is also very common, with some exceptions, to treat these various pooled representation in a completely identical fashion, e.g.\ by comparing them all with cosine similarity.
Intuitively, however, we suggest that the statistics of $W_{\text{pool}}$ must heavily depend on the pooling function $f_\text{{pool}}$ and thus each such pooled random variable should be studied in its own right.
To illustrate this point, we would like to reveal the very different nature of mean- and max- and min-pooled sentence vectors though a practical example.

\subsection{Statistics of the Pooled Representations: A Practical Analysis}
\label{subsec:practice}

\begin{figure*}[h]
	\centering
	\includegraphics[width=\textwidth]{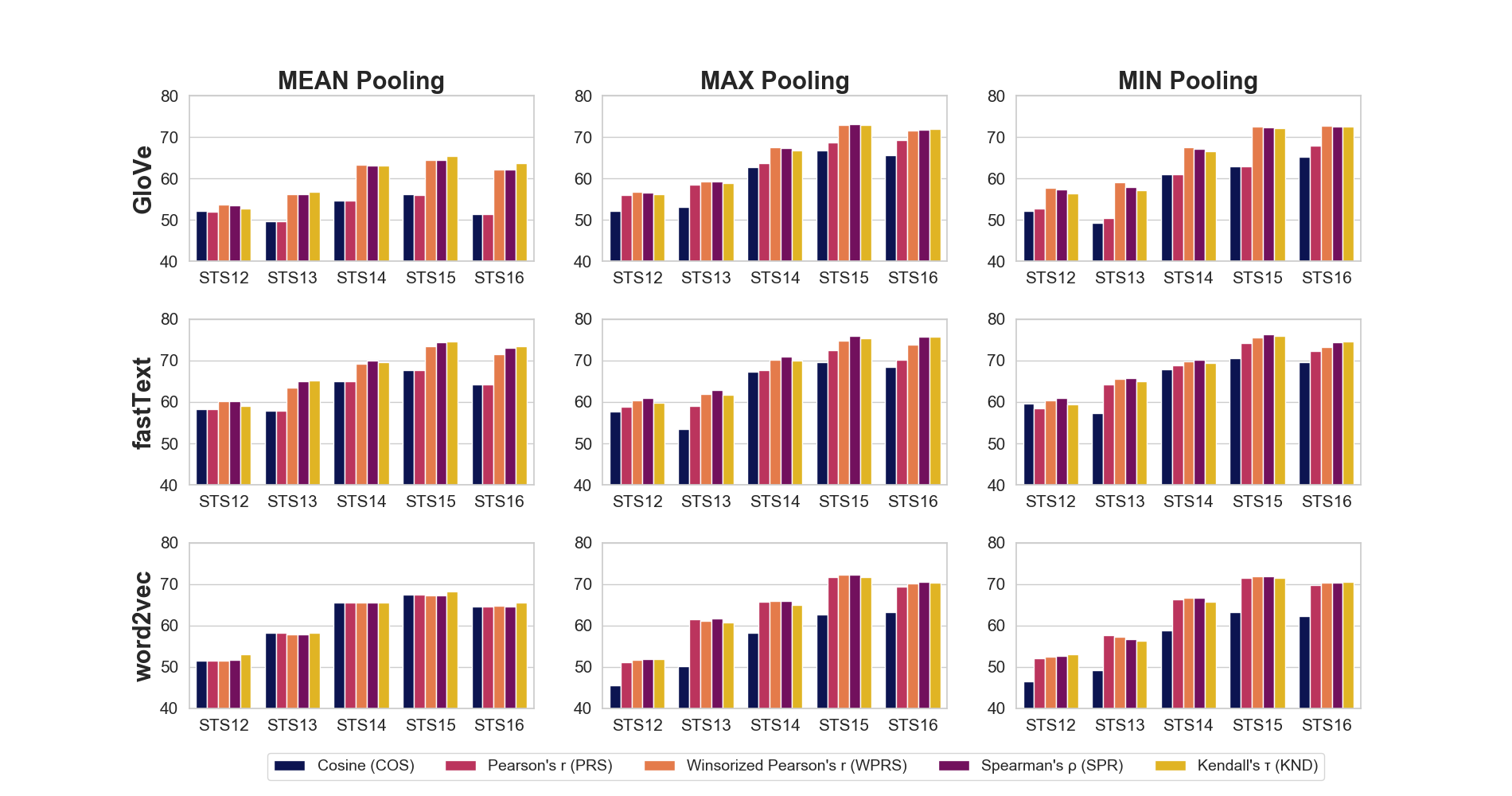}
	\caption{
	Bar plots of Pearson correlation on STS tasks between human scores and the following set-based similarity metrics: Cosine similarity (COS), Pearson's $r$ (PRS), Winsorized Pearson's $r$ (WPRS), Spearman's $\rho$~(SPR), and Kendall's $\tau$~(KND). Plots generated for three pooling methods and the following word embedding models:
	GloVe~\cite{Pennington2014}, fastText~\cite{Bojanowski2016}, and word2vec~\cite{Mikolov2013a,Mikolov2013b}.
	}
	\label{fig:sts-bars}
\end{figure*}

\begin{figure*}[h]
	\centering
	\includegraphics[width=\textwidth]{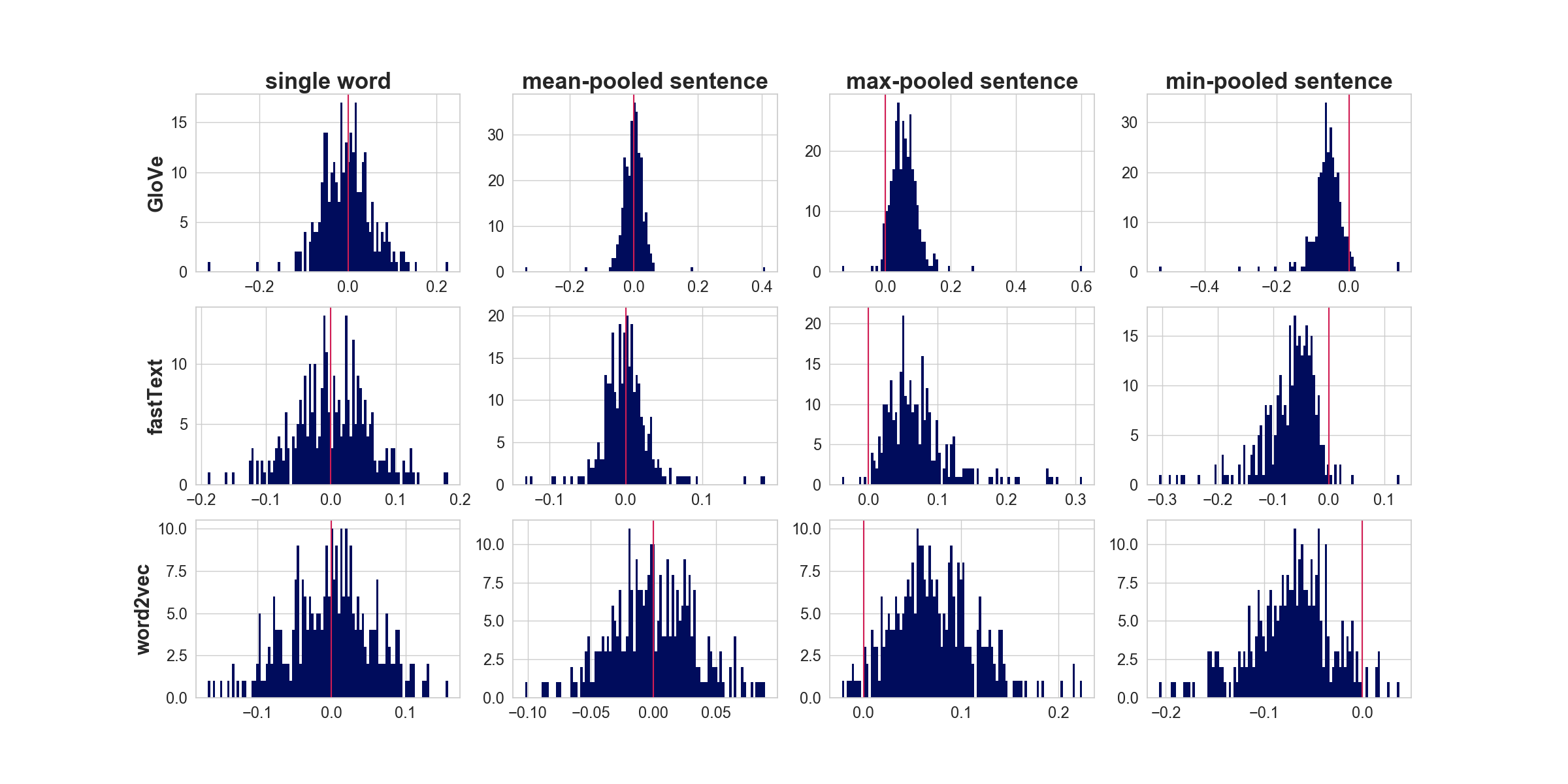}
	\caption{
	Histograms for word embeddings of the word ``cats'' and pooled representations of the embeddings for the words in the sentence ``I like cats because they are very cute animals''. 
	}
	\label{fig:sentence-hist}
\end{figure*}

Let us begin by examining sentence vectors obtained through mean-pooling. Recall that for common word embedding models, the mean across 300 dimensions of a single word embedding $\mbw_{(i)}$ happens to be close to zero (relative to the dimensions).
By the linearity of expectation, we have that ${\mathbb E}[W_{\text{mean}}] = {\mathbb E}\left[\sum_{i=1}^{k} W_i\right] = \sum_{i=1}^{k} {\mathbb E}\left[W_i\right]$, and so the mean across $\mbw_{\text{mean}}$ will also be close to zero at least for small $k$.
In practice, this seems to hold even for moderate $k$ in naturally occurring sentences, as seen in \Cref{fig:means}.
Based on this, we expect Pearson correlation and cosine similarity to have almost identical performance on the downstream tasks, which is confirmed in \Cref{fig:sts-bars}.

On the other hand, intuition tells us that the means of the max-pooled vectors will be shifted to the right because of the max operation, which we see in \Cref{fig:means}.
In this case, cosine similarity and Pearson correlation will yield different results and, in fact, Pearson's $r$ considerably outperforms cosine on the downstream tasks (\Cref{fig:sts-bars}).
This in turn empirically adds weight to the statistical interpretation (correlation) over its geometrical counterpart (angle between vectors).

Recall also that unlike word2vec, GloVe and fastText vectors feature heavy univariate outliers, and the same can be expected to hold for the pooled representations; an example is shown in \Cref{fig:sentence-hist}.
In case of mean-pooled vectors, this particular departure from normality can be successfully detected by the Shapiro-Wilk normality test, informing the appropriate choice of the correlation coefficient (Pearson's $r$ or robust rank correlation).
By contrast, such procedure cannot be readily applied to max-pooled and min-pooled vectors as by construction they exhibit additional departures from normality, such as positive and negative skew respectively.
It is always a good idea to consult visualisations for such vectors, such as the ones in \Cref{fig:sentence-hist}.
Interestingly though, we do observe the some noteworthy regularities, which we describe further in \Cref{sec:experiments}.

The above example is meant to illustrate that even the simplest pooled random variables show strikingly different statistics depending on the aggregation.
While the abundance of various pooling operations may be intimidating, the resulting vectors are always subject to the many tools of univariate statistics.
As we hope to have shown, even crude analysis can shed light on the nature of these textual representations, which in turn has notable practical implications,
as we will see in \Cref{sec:experiments}.

\subsection{Correlations between Random Vectors}
\label{subsec:correlationsBetweenRandomVectors}

Exactly as before, suppose we have two sentences $S_1 = \{\mbx_{(1)}, \mbx_{(2)}, \ldots, \mbx_{(k)}\}$ and $S_2 = \{\mby_{(1)}, \mby_{(2)}, \ldots, \mby_{(l)}\}$ and the corresponding random vectors $\mbX = (X_1, X_2, \ldots, X_k)$ and $\mbY = (Y_1, Y_2, \ldots, Y_l)$.
At this point it is important to emphasise again that we relate each word vector $\mbx_{i}$ to a random variable $X_i$ and treat the dimensions of $\mbx_{i}$ as $D$ observations from that variable, and similarly for $\mby_{i}$ and $Y_i$.
In contrast with the pooling-based approaches, our task here is to find a suitable correlation coefficient directly between the random vectors $\mbX$ and $\mbY$.
We begin by recalling the expression for the basic univariate Pearson's $r$:

\begin{equation}
r_{XY} = \frac{\mathbb{E}_{X, Y}\left[(X - \mu_X)(Y - \mu_Y)\right]}{\sigma_X \sigma_Y},
\end{equation}
where
\[
    \mu_X = \mathbb{E}[X],\quad \sigma_X = \sqrt{\mathbb{E}\left[X^2\right]-\mu_X^2},
\]
and similarly for $\mu_Y$ and $\sigma_Y$.
The covariance term $\text{cov}(X, Y)$ in the numerator is readily generalised to random vectors by the following cross-covariance operator between reproducing kernel Hilbert spaces (RKHS) $\mathcal{F}$ and $\mathcal{G}$
\begin{equation}
\label{eq:cross_cov_op}
\mathcal{C}_{\mbX\mbY} = \mathbb{E}_{\mbX,\mbY}\left[\left(\phi(\mbX) - \mu_{\mbX}\right)\otimes\left(\psi(\mbY) - \mu_{\mbY}\right)\right],
\end{equation}
where $\otimes$ denotes the tensor product and $\mu_{\mbX} = \mathbb{E}_{\mbX}\left[\phi(\mbX)\right]$, $\mu_{\mbY} = \mathbb{E}_{\mbY}\left[\psi(\mbY)\right]$.
Here $\phi$ and $\psi$ are the feature maps such that $\langle \phi(\mbx), \phi(\mbx') \rangle_\mathcal{F} = K(\mbx, \mbx')$ and $\langle \psi(\mby), \psi(\mby') \rangle_\mathcal{G} = L(\mby, \mby')$, where $K$ and $L$ are the kernels associated with RKHS $\mathcal{F}$ and $\mathcal{G}$ respectively.
Note that if $\phi$ and $\psi$ are the identity maps, the cross-covariance operator~\eqref{eq:cross_cov_op} simply becomes the cross-covariance matrix
\begin{equation*}
\label{eq:cross_cov_mat}
\mb{C}_{\mbX\mbY} = \mathbb{E}_{\mbX,\mbY}\left[\left(\mbX - \mu_{\mbX}\right)\left(\mbY - \mu_{\mbY}\right)^{T}\right].
\end{equation*}
\citet{gretton2005measuring} define the Hilbert-Schmidt independence criterion (HSIC) to be the squared Hilbert-Schmidt norm $||\mathcal{C}_{\mbX\mbY}||^2_{\text{HS}}$ of~\eqref{eq:cross_cov_op} and derive an expression for it in terms of kernels $K$ and $L$
\begin{align*}
\text{HSIC}(\mbX, \mbY, K, L) = \\
\mathbb{E}_{\mbX,\mbX',\mbY,\mbY'}\left[K(\mbX, \mbX')L(\mbY, \mbY')\right] \\
+ \mathbb{E}_{\mbX,\mbX'}\left[K(\mbX, \mbX')\right]\mathbb{E}_{\mbY,\mbY'}\left[L(\mbY, \mbY')\right] \\
- 2\mathbb{E}_{\mbX,\mbY}\left[ \mathbb{E}_{\mbX'}\left[K(\mbX, \mbX')\right]\mathbb{E}_{\mbY'}\left[L(\mbY, \mbY')\right] \right].
\end{align*}
They also show the empirical estimate of it to be
\begin{equation}
\label{eq:hsic}
\text{HSIC}(\mbK, \mbL) = (D-1)^{-2}\Tr (\mbK \mbH \mbL \mbH),
\end{equation}
where $\mbH = \mbI - \frac{1}{D}\mathbf{1}\mathbf{1}^T$ is the centering matrix and $\mbK = K(\mbX^{(i)}, \mbX^{(j)}), \mbL = L(\mbY^{(i)}, \mbY^{(j)}), i,j = 1{:}D$ are the kernel (Gram) matrices of observations.
Crucially, the kernel evaluations for $\mbK$ take place between $\mbX^{(i)} = (\mbx^{i}_{(1)}, \mbx^{i}_{(2)}, \ldots, \mbx^{i}_{(k)})$ and $\mbX^{(j)} = (\mbx^{j}_{(1)}, \mbx^{j}_{(2)}, \ldots, \mbx^{j}_{(k)})$
and not between the individual word embeddings $\mbx_{(i)}$ and $\mbx_{(j)}$, and similarly for $\mbL$.
Thus, both $\mbK$ and $\mbL$ are square matrices of dimension $D \times D$ .
Indeed, for~\eqref{eq:hsic} to make sense, the dimensions of $\mbK$ and $\mbL$ must match.
The matching dimension in our case is the word embedding dimension $D$, while the number of words $k$ and $l$ in the sentences may vary.
This is in line with our formalism, which models word vectors as random variables and their dimensions as observations.

Finally, the Centered Kernel Alignment (CKA) \citep{cortes2012} is simply defined as
\begin{equation}
\label{eq:cka}
\text{CKA}(\mbK, \mbL) = \frac{\text{HSIC}(\mbK, \mbL)}{\sqrt{\text{HSIC}(\mbK, \mbK)} \sqrt{\text{HSIC}(\mbL, \mbL)}}.
\end{equation}
We see now that CKA not only generalises the squared Pearson correlation to the multivariate case, it also allows it to operate in high-dimensional feature spaces, as commonly done in the kernel literature.
The reason this is useful is that under certain conditions (when $K$ and $L$ are characteristic kernels), HSIC can detect any existing dependence with high probability, as the sample size increases \citep{gretton2005kernel}.
One can also consider the Uncentered Kernel Alignment (or simply KA) \citep{cristianini2002}, which can then be seen as a similar generalisation but for the univariate cosine similarity.
To the best of our knowledge, KA and CKA in general have never been applied before to measure semantic similarity between sets of word embeddings; therefore this work seeks to introduce them as standard definitions for squared Pearson's $r$ and cosine similarity for such sets.

\begin{table}[t]
	\centering
	\begin{tabular}{ksssss}
		\toprule
		\textbf{Approach~~~~STS} & \textbf{12~~} & \textbf{13~~} & \textbf{14~~} & \textbf{15~~} & \textbf{16~~} \\
		\midrule
		\midrule
		\multicolumn{6}{l}{\small\textit{Deep Learning Approaches}}\\
		\midrule
		ELMo (BoW)        & 55~~          & 53~~          & 63~~          & 68~~          & 60~~          \\
		Skip-Thought      & 41~~          & 29~~          & 40~~          & 46~~          & 52~~          \\
		InferSent         & \textbf{61~~}          & 56~~          & 68~~          & 71~~          & 71~~          \\
		USE (DAN)         & 59~~          & 59~~          & 68~~          & 72~~          & 70~~          \\
		USE (Transformer) & \textbf{61~~}          & 64~~          & 71~~          & 74~~          & 74~~          \\
		STN (multitask)   & 60.6        & 54.7\textsuperscript{\textdagger}        & 65.8        & 74.2        & 66.4        \\
		BERT-Base & 46.9 & 52.8 & 57.2 & 63.5 & 64.5\\  
		BERT-Large & 42.6 & 47.4 & 49.3 & 55.6 & 60.4\\  
		\midrule
		\multicolumn{6}{l}{\small\textit{Set-based Approaches}}\\
		\midrule
		WMD & 53.0 & 45.9 & 57.2 & 66.9 & 63.1\\  
		SoftCardinality & 55.9 & 50.5 & 59.0 & 66.2 & 65.1\\  
		DynaMax    & 60.9 &  60.3 & 69.5 & \textbf{76.7} & 74.6 \\
		SIF+PCA & 58.6 & \textbf{67.3} & 70.5 & 73.5 & 71.7\\  
		MeanPool+COS    & 58.3 & 57.9 & 64.9 & 67.6 & 64.3 \\
		MaxPool+COS & 57.7 & 53.5 & 67.2 & 69.5 & 68.5\\  
		MeanPool+SPR     & 60.2        & 65.1        & 70.1        & 74.4        & 73.0          \\
		MaxPool+SPR\textsuperscript{\MineSign} & \textbf{61.0} & 62.9 & 70.9 & 75.9 & \textbf{75.8}\\  
		CKA Linear\textsuperscript{\MineSign} & 59.8 & 62.1 & 69.5 & 74.6 & 70.3\\  
		CKA Gaussian\textsuperscript{\MineSign} & 60.5 & 63.8 & \textbf{71.6} & 76.3 & 73.7\\  
		CKA dCor\textsuperscript{\MineSign} & \textbf{61.0} & 63.2 & 71.5 & 75.6 & 72.4\\  
		\bottomrule
	\end{tabular}
	\caption{
		Mean Pearson correlation on STS tasks for Deep Learning and Set-based methods using fastText.
		Methods proposed in this work are denoted with \MineSign. 
		Values in bold indicate best results per task. 
		Previous results are taken from \citet{perone2018evaluation}, \citet{Subramanian2018} and \citet{zhelezniak2019a,zhelezniak2019b}.
		\textsuperscript{\textdagger}~indicates the only STS13 result (to our knowledge) that includes the SMT subtask.
	}
	\label{tab:sts-sota}
\end{table}

\section{Experiments}
\label{sec:experiments}

We now empirically demonstrate the power of the methods and statistical analysis presented in \Cref{sec:theory}, through
a set of evaluations on the Semantic Textual Similarity (STS) tasks series 2012-2016~\citep{Agirre2012, Agirre2013a, Agirre2014, Agirre2015, Agirre2016, Cer2017}.
For methods involving pretrained word embeddings, we use fastText \citep{Bojanowski2016} trained on Common Crawl (600B tokens),
as previous evaluations have indicated that fastText vectors have uniformly the best performance on these tasks out of commonly-used pretrained unsupervised word vectors
\citep{Conneau2017,perone2018evaluation,zhelezniak2019b,zhelezniak2019a}.
We provide experiments and significance analysis for additional word vector in the Appendix.
The success metric for the STS tasks is the Pearson correlation between the sentence similarity scores provided by human annotators and the scores generated by a candidate algorithm.
Note that the dataset for the STS13 SMT subtask is no longer publicly available, so the mean Pearson correlation for STS13 reported in our experiments has been re-calculated accordingly.
The code for our experiments builds on the SentEval toolkit \citep{conneau2018senteval} and is available on GitHub\footnote{\url{https://github.com/Babylonpartners/corrsim}}.

We first conduct a set of experiments to validate the observations of \Cref{subsec:practice,subsec:correlationsBetweenPooledVariables} regarding
the performance of cosine similarity and various univariate correlation coefficients when applied to pooled word vectors.
These results are depicted in \Cref{fig:sts-bars}, for which we can make the following observations.

First, max and min-pooled vectors consistently outperform mean-pooled vectors when all three representations are compared with Pearson correlation.
We hypothesise that this is in part because max and min-pooling remove the outliers (to which Pearson's $r$ is very sensitive) from at least one tail of the distribution whereas mean-pooled vectors have outliers in both tails.
This outlier-removing property, however, cannot be taken as a sole explanation behind excellent performance of max-pooled vectors,
as max-pooling still tends to outperform mean-pooling when both are compared with correlations that are robust to outliers, as well as on word vectors that have very few outliers to begin with (e.g. word2vec).

In addition, the strong performance of rank correlation coefficients (Spearman's~$\rho$ and Kendall's~$\tau$) comes solely from their robustness to outliers, as clipping (winsorizing) the top and bottom 5\% of the values and then proceeding with Pearson's $r$ closes the gap almost completely.
Consistently, on vectors with few outliers (word2vec), Pearson's $r$ achieves the same performance as rank correlations even without winsorization.
However, unlike outliers, positive (negative) skew of max- (min-) pooled vectors does not seem to hurt Pearson's $r$ on STS tasks.

Next, we conduct evaluations of the methods proposed in this work alongside other deep learning and set-based similarity measures for STS from the literature.
The methods we compare are as follows:
\begin{itemize}
    \item \emph{Deep representation approaches:} BoW with ELMo embeddings \citep{Peters2018}, Skip-Thought \citep{Kiros2015},
    InferSent \citep{Conneau2017}, Universal Sentence Encoder both DAN and Transformer \citep{USE2018}, STN multitask embeddings \citep{Subramanian2018},
    and BERT 12- and 24-layer models \citep{devlin2018bert}.
    \item \emph{Set-based similarity measures:} Word Mover's Distance (WMD) \citep{Kusner2015}, soft-cardinality with Jaccard coefficient \citep{Jimenez2012},
    DynaMax with Jaccard \citep{zhelezniak2019a}, mean- and max-pooled word vectors with cosine similarity (COS), and mean-pooled word vectors with Spearman correlation (SPR) \citep{zhelezniak2019b}.
    \item \emph{Proposed set-based approaches:} max-pooled word vectors with Spearman correlation, CKA with linear kernel (also known as RV-coefficient),
    CKA with Gaussian kernel (median estimation for $\sigma^2$), and CKA with distance kernel (distance correlation).
\end{itemize}
Note that for BERT we evaluated all pooling strategies available in bert-as-service \citep{xiao2018bertservice} applied to either the last or second-to-last layers and report
results for the best-performing combination, which was mean-pooling on the last layer for both model sizes.
\begin{table}[t]
	\centering
	\begin{tabular}{kk}
		\toprule
		\textbf{Approach} & \textbf{Time complexity} \\
		\midrule
        MaxPool+SPR & $O(nd + d\log d)$ \\
		CKA & $O(nd^2 + d^2)$ \\
        DynaMax & $O(n^2d)$ \\
		SoftCard & $O(n^2d)$ \\
        WMD & $O(n^3\log n\cdot d)$ \\
        WMD (relaxed) & $O(n^2d)$ \\
		\bottomrule
	\end{tabular}
	\caption{
	Computational complexity of some of the set-based STS methods discussed in this paper.
    Here $n$ is the sentence length and $d$ is the dimensionality of the word embeddings.
	}
	\label{tab:complexity}
\end{table}

Our results are presented in \Cref{tab:sts-sota}.
We can clearly see that deep learning-based methods do not shine on STS tasks, while simple compositions of word vectors
can perform extremely well, especially when an appropriate correlation coefficient is used as the similarity measure.
Indeed, the performance of max-pooled vectors with Spearman correlation approaches or exceeds that of more expensive or
offline methods like that of \citet{Arora2017}, which performs PCA computations on the entire test set.
Additionally, while the multivariate correlation methods such as CKA are more computationally expensive than pooling-based approaches (see \Cref{tab:complexity}), they can
provide performance boost on some tasks, making the cost worth it depending on the application.
Finally, we conducted an exploratory error analysis and found that many errors are due to the well-known inherent weaknesses of word embeddings.
For example, the proposed approaches heavily overestimate similarity when two sentences contain antonyms or when one sentence is the negation of the other.
We illustrate these and other cases in the Appendix.

\section{Conclusion}
\label{sec:conclusion}

In this work we investigate the application of statistical correlation coefficients to sets of word vectors as a method
for computing semantic textual similarity (STS).
This can be done either by pooling these word vectors and computing univariate correlations between the resulting
representations, or by applying multivariate correlation coefficients to the sets of vectors directly.

We provide further empirical evidence that outliers in word vector distributions disrupt performance of set-based similarity metrics as previously shown~\cite{zhelezniak2019b}.
We also show working methods for solving or avoiding the issue through vector pooling operations, robust correlations or winsorization.
In addition, we found that pooling operations in conjunction with univariate correlation coefficients yield one of the strongest results on downstream STS tasks, while being computationally much more efficient than competing set-based methods.
Our findings are supported by a combination of statistical analysis, practical examples and visualisations, and empirical evaluation on standard benchmark datasets.

Both proposed families of approaches serve as strong baselines for future research into STS, as well as useful
algorithms for the practitioner, being efficient and simple to implement.

We believe our findings speak to the efficacy of the statistical perspective on word embeddings, which we hope will encourage
others to explore further implications of not only  this particular framework, but also completely novel interpretations of
textual representations.

\section*{Acknowledgements}
\label{sec:acknowledgements}

We would like to thank the three anonymous reviewers for their useful feedback and suggestions.

\bibliography{emnlp2019}
\bibliographystyle{acl_natbib}
\clearpage

\appendix
{\bf \noindent \large Appendix}

\section{Error Analysis}\label{sec:errorAnalysis}
While the proposed approaches enjoy strong performance on the STS benchmarks relative to the competing methods, the Pearson correlations between gold and system scores remain consistently below 0.9 in all subtasks.
It would be extremely useful to establish which similarities are not captured very well by these approaches, at least as judged by humans on the 0 to 5 scale established in ~\citep{Agirre2012}.
For concreteness, we limit our exposition to MaxPool Spearman, noting that similar conclusions hold for CKA-based methods too.

First, we linearly transform the system scores into the range $[0, 5]$, thus making them comparable to gold scores while preserving Pearson correlation.
Then, in each subtask we select 5 sentence pairs with the largest absolute difference between the gold and the system score.
After that, we manually examine the obtained dataset, focusing predominantly on shorter sentences, where the errors are often obvious and easy to explain.
Even under these restrictions, we can readily distinguish between 5 different types of errors, summarised in \Cref{tab:error_analysis}.
On the one hand, the system heavily underestimates the similarity score when two sentences use completely different vocabulary yet have identical meaning (Type \RomanNumeralCaps{1}).
On the other hand, it tends to overestimate the similarity when the sentences use very related or even the same words but have different meaning (Types \RomanNumeralCaps{2} \& \RomanNumeralCaps{3}).
The similarity is also overestimated when two sentences contain antonyms, or when one sentence is a negation of the other (Types \RomanNumeralCaps{4} \& \RomanNumeralCaps{5} respectively).
A lot of these flaws can be traced back to the well-known weaknesses of word embeddings and the distributional hypothesis, such as mixing together semantic similarity and conceptual relatedness \citep{Hill2015, Mrksic2016}, failure to distinguish synonyms from antonyms \citep{Mohammad2008, Mrksic2016} and problems with negation.
We hope that any counter-measures to these weaknesses will also improve the proposed sentence-level systems.

\section{Significance Analysis}\label{sec:sigAnalysis}

Following the procedure described in \citet{zhelezniak2019a}, we construct 95\% BCa confidence intervals for the delta in performance between two systems.
The key results are as follows.
MaxPool Spearman overall statistically outperforms DynaMax \citep{zhelezniak2019a} when word vectors are highly non-normal (GloVe) and looses when word vectors seem mostly normal (word2vec), which is in line with our main discussion.
Next, max-pooling outperforms mean-pooling on the majority of subtasks for all word vector models.
Finally, MaxPool Spearman is overall comparable to CKA Gaussian, with the exception of word2vec where CKA is slightly better.

\clearpage

\begin{table*}[t!]
\setlength{\tabcolsep}{3pt}
\small
\centering
\begin{tabular} { c p{6.4cm}  p{5.6cm} c c c } \toprule
Type & Sentence 1 & Sentence 2 &  Gold & Sys. & $\Delta$ \\\midrule
\multirow{7}{*}{\RomanNumeralCaps{1}} & \textit{Identical meaning but different words} & & & &   \\\cmidrule{2-3}
& restrict or confine & place limits on (extent or access). & 4.75 & 1.69 & +3.06 \\
& the reduction of the extent of something, e.g, its size, importance or quantity & change toward something smaller or lower. & 4.4 & 1.49 & +2.91 \\
& an occasion on which people can assemble for social interaction and entertainment. & festive social event, celebration & 4.25 & 1.33 & +2.92 \\

\midrule

\multirow{4}{*}{\RomanNumeralCaps{2}} & \textit{Related words but different meaning} & & & &   \\\cmidrule{2-3}
& a man is playing the piano. & a woman is playing the violin. & 1 & 3.66 & -2.66 \\
& indonesian president to visit uk & indonesian president to visit australia & 1.4 &  4.29 &  -2.89 \\
& a grey, black, and white cat looking at the camera. & a black and white dog looking at the camera. & 1 & 3.95 & -2.95 \\

\midrule

\multirow{3}{*}{\RomanNumeralCaps{3}} & \textit{Same keywords but different meaning} & & & &   \\\cmidrule{2-3}
& why do you need to peel peaches to can them? & how to peel peaches? & 1 &  4.57 &  -3.57 \\
& what does it mean to write a song in a certain key? & is it possible to write a song without a key? & 1 & 4.04 & -3.04 \\

\midrule

\multirow{4}{*}{\RomanNumeralCaps{4}} & \textit{Antonyms} & & & &   \\\cmidrule{2-3}
& chinese stocks close higher midday friday & chinese stocks open lower friday & 1 & 3.9 & -2.9 \\
& the act of beginning something new. & the act of ending something. & 0.8 &  3.67 & -2.87 \\
& higher than per cent but not very high. & lower than per cent but not very low. & 1 & 4.22 & -3.22 \\

\midrule

\multirow{4}{*}{\RomanNumeralCaps{5}} & \textit{Negation} & & & &   \\\cmidrule{2-3}
& you are a christian. & therefore you are not a christian. & 1.4 & 4.38 & -2.98 \\
& you should do it. & you should never do it. & 1 &  4.56 &  -3.56 \\
& it's not a good idea. & it's a good idea to do both. & 1 & 3.9 & -2.9 \\

\bottomrule
\end{tabular}
\caption{
Error analysis for MaxPool Spearman.
Each entry contains a sentence pair, the gold similarity score, the scaled system similarity score, and the difference between the two scores.
Errors are categorised into 5 types.
The system heavily underestimates the similarity score when two sentences use different vocabulary yet have identical meaning (Type \RomanNumeralCaps{1}).
Inversely, it overestimates the similarity when the sentences use very related or even the same words but have different meaning (Types \RomanNumeralCaps{2} \& \RomanNumeralCaps{3}).
The similarity is also overestimated when two sentences contain antonyms, or when one sentence is a negation of the other (Types \RomanNumeralCaps{4} \& \RomanNumeralCaps{5} respectively).
}
\label{tab:error_analysis}
\end{table*}

\onecolumn
\begin{sidewaystable}[h]
\centering
\begin{tabular}{s@{\hskip 7pt}k@{\hskip 9pt}ssk@{\hskip 9pt}ssk@{\hskip 9pt}ssk}
\toprule
 &  & \multicolumn{3}{c}{\textbf{GloVe}} & \multicolumn{3}{c}{\textbf{fastText}} & \multicolumn{3}{c}{\textbf{word2vec}}\\
\cline{3-11}
 &  & \textbf{MPS} & \textbf{DMX} & \textbf{$\Delta$95\% CI} & \textbf{MPS} & \textbf{DMX} & \textbf{$\Delta$95\% CI} & \textbf{MPS} & \textbf{DMX} & \textbf{$\Delta$95\% CI}\\
\midrule
\toprule
\textbf{\multirow{5}{*}{\rotatebox[origin=c]{90}{STS12}}} & MSRpar & 40.00 & \textbf{49.41} & [-12.75, -6.26] & 44.48 & \textbf{48.94} & [-7.48, -1.53] & 36.70 & \textbf{41.74} & [-7.74, -2.40]\\
 & MSRvid & \textbf{77.66} & 71.92 & [4.35, 7.27] & \textbf{82.44} & 76.20 & [5.00, 7.59] & 74.34 & \textbf{76.86} & [-3.65, -1.47]\\
 & SMTeuroparl & \textbf{46.52} & \textbf{48.43} & [-4.61, 0.84] & 50.18 & \textbf{53.08} & [-5.44, -0.50] & \textbf{34.13} & 28.03 & [3.63, 8.38]\\
 & surprise.OnWN & \textbf{69.23} & \textbf{69.86} & [-2.21, 0.93] & \textbf{73.12} & \textbf{72.79} & [-1.01, 1.70] & 69.06 & \textbf{71.26} & [-3.38, -1.04]\\
 & surprise.SMTnews & \textbf{49.28} & \textbf{51.47} & [-5.37, 1.40] & \textbf{55.01} & \textbf{53.26} & [-1.72, 5.61] & 45.09 & \textbf{50.44} & [-8.09, -2.78]\\
\midrule
\textbf{\multirow{3}{*}{\rotatebox[origin=c]{90}{STS13}}} & FNWN & \textbf{46.16} & \textbf{39.79} & [-2.72, 15.43] & \textbf{44.14} & \textbf{42.34} & [-7.33, 11.19] & \textbf{49.66} & \textbf{42.34} & [-0.97, 17.12]\\
 & headlines & \textbf{70.60} & \textbf{69.91} & [-0.75, 2.10] & \textbf{73.04} & \textbf{73.13} & [-1.26, 1.04] & \textbf{65.89} & \textbf{66.66} & [-1.97, 0.44]\\
 & OnWN & \textbf{61.03} & 52.12 & [6.50, 11.66] & \textbf{71.37} & 65.35 & [3.97, 8.24] & \textbf{69.40} & \textbf{69.36} & [-1.23, 1.37]\\
\midrule
\textbf{\multirow{6}{*}{\rotatebox[origin=c]{90}{STS14}}} & deft-forum & \textbf{44.33} & \textbf{43.29} & [-2.29, 4.56] & \textbf{52.50} & 47.16 & [2.01, 8.80] & \textbf{45.60} & \textbf{47.27} & [-4.51, 1.22]\\
 & deft-news & \textbf{70.69} & \textbf{70.55} & [-2.62, 2.82] & \textbf{70.64} & \textbf{71.04} & [-3.01, 2.01] & 62.84 & \textbf{65.84} & [-5.30, -0.82]\\
 & headlines & \textbf{65.65} & \textbf{64.49} & [-0.44, 2.76] & \textbf{68.38} & \textbf{68.22} & [-1.09, 1.39] & 62.00 & \textbf{63.66} & [-3.03, -0.27]\\
 & images & \textbf{78.98} & 75.05 & [2.44, 5.51] & \textbf{81.46} & 79.39 & [0.95, 3.24] & 78.33 & \textbf{80.51} & [-3.28, -1.14]\\
 & OnWN & \textbf{69.20} & 63.00 & [4.46, 8.07] & \textbf{75.92} & 72.83 & [1.86, 4.44] & \textbf{75.35} & \textbf{75.43} & [-0.90, 0.80]\\
 & tweet-news & \textbf{74.77} & \textbf{74.30} & [-1.49, 2.58] & 76.38 & \textbf{78.41} & [-3.39, -0.70] & 71.17 & \textbf{75.47} & [-5.70, -2.96]\\
\midrule
\textbf{\multirow{5}{*}{\rotatebox[origin=c]{90}{STS15}}} & answers-forums & \textbf{67.33} & 61.94 & [1.75, 9.45] & 69.89 & \textbf{73.57} & [-6.98, -0.42] & 62.27 & \textbf{66.44} & [-7.59, -0.83]\\
 & answers-students & 71.28 & \textbf{73.53} & [-3.84, -0.71] & 73.30 & \textbf{75.82} & [-3.97, -1.12] & \textbf{74.20} & \textbf{75.07} & [-2.08, 0.25]\\
 & belief & \textbf{72.83} & 67.21 & [2.28, 9.46] & \textbf{76.67} & \textbf{76.14} & [-2.21, 3.41] & \textbf{74.17} & \textbf{75.83} & [-3.67, 0.55]\\
 & headlines & \textbf{72.14} & \textbf{72.26} & [-1.41, 1.08] & \textbf{74.78} & \textbf{74.45} & [-0.71, 1.31] & 68.49 & \textbf{69.95} & [-2.49, -0.46]\\
 & images & \textbf{81.83} & 79.30 & [1.15, 3.98] & \textbf{84.84} & 83.33 & [0.50, 2.50] & 82.60 & \textbf{83.80} & [-2.08, -0.38]\\
\midrule
\textbf{\multirow{5}{*}{\rotatebox[origin=c]{90}{STS16}}} & answer-answer & \textbf{61.71} & \textbf{59.72} & [-1.64, 5.58] & \textbf{66.58} & \textbf{63.30} & [-1.24, 8.22] & \textbf{58.65} & \textbf{58.78} & [-3.09, 2.90]\\
 & headlines & \textbf{70.57} & \textbf{71.71} & [-3.04, 0.80] & \textbf{72.81} & \textbf{73.40} & [-2.50, 1.10] & \textbf{67.87} & \textbf{68.18} & [-1.97, 1.38]\\
 & plagiarism & \textbf{78.02} & \textbf{79.92} & [-4.41, 0.44] & \textbf{82.97} & \textbf{82.68} & [-1.61, 2.22] & \textbf{80.59} & \textbf{82.05} & [-3.11, 0.02]\\
 & postediting & \textbf{81.11} & \textbf{80.48} & [-1.30, 2.97] & \textbf{82.31} & \textbf{84.15} & [-3.73, 0.15] & 78.97 & \textbf{81.73} & [-4.91, -1.13]\\
 & question-question & \textbf{66.88} & \textbf{63.51} & [-0.58, 7.85] & \textbf{74.19} & 69.71 & [1.03, 8.16] & \textbf{66.79} & \textbf{65.74} & [-2.99, 4.85]\\
\midrule
\end{tabular}
\captionsetup{width=6in}
\caption{\textbf{MaxPool Spearman vs. DynaMax:}
	Pearson correlations between human sentence similarity scores and a generated scores.
    Values in bold represent the best result for a subtask given a set of word vectors,
	based on a 95\% BCa confidence interval~\cite{Efron1987} on the differences between the two correlations.
    In cases of no significant difference, both values are in bold.}
\label{tab:mps-v-dmx}
\end{sidewaystable}
\begin{sidewaystable}
\centering
\begin{tabular}{s@{\hskip 7pt}k@{\hskip 9pt}ssk@{\hskip 9pt}ssk@{\hskip 9pt}ssk}
\toprule
 &  & \multicolumn{3}{c}{\textbf{GloVe}} & \multicolumn{3}{c}{\textbf{fastText}} & \multicolumn{3}{c}{\textbf{word2vec}}\\
\cline{3-11}
 &  & \textbf{MPS} & \textbf{ASP} & \textbf{$\Delta$95\% CI} & \textbf{MPS} & \textbf{ASP} & \textbf{$\Delta$95\% CI} & \textbf{MPS} & \textbf{ASP} & \textbf{$\Delta$95\% CI}\\
\midrule
\toprule
\textbf{\multirow{5}{*}{\rotatebox[origin=c]{90}{STS12}}} & MSRpar & \textbf{40.00} & \textbf{35.90} & [-1.09, 9.00] & \textbf{44.48} & \textbf{39.66} & [-0.17, 9.82] & \textbf{36.70} & \textbf{38.79} & [-6.36, 2.07]\\
 & MSRvid & \textbf{77.66} & 68.80 & [6.87, 11.04] & \textbf{82.44} & 81.02 & [0.22, 2.68] & 74.34 & \textbf{77.88} & [-4.98, -2.20]\\
 & SMTeuroparl & \textbf{46.52} & \textbf{48.73} & [-5.82, 1.45] & \textbf{50.18} & \textbf{50.29} & [-3.85, 3.41] & \textbf{34.13} & 16.96 & [12.54, 21.09]\\
 & surprise.OnWN & \textbf{69.23} & 66.66 & [0.12, 5.28] & \textbf{73.12} & \textbf{73.15} & [-1.79, 1.88] & \textbf{69.06} & \textbf{70.75} & [-3.64, 0.15]\\
 & surprise.SMTnews & \textbf{49.28} & \textbf{47.12} & [-3.49, 7.65] & \textbf{55.01} & \textbf{56.67} & [-6.13, 3.14] & 45.09 & \textbf{53.93} & [-13.49, -4.14]\\
\midrule
\textbf{\multirow{3}{*}{\rotatebox[origin=c]{90}{STS13}}} & FNWN & \textbf{46.16} & \textbf{43.21} & [-8.69, 14.23] & \textbf{44.14} & \textbf{49.40} & [-14.85, 3.81] & \textbf{49.66} & \textbf{40.73} & [-0.61, 19.27]\\
 & headlines & \textbf{70.60} & 67.59 & [1.16, 4.86] & \textbf{73.04} & \textbf{71.53} & [-0.12, 3.15] & \textbf{65.89} & \textbf{65.48} & [-1.29, 2.19]\\
 & OnWN & \textbf{61.03} & 57.66 & [0.39, 6.44] & 71.37 & \textbf{74.33} & [-5.17, -1.00] & \textbf{69.40} & \textbf{67.49} & [-0.24, 4.08]\\
\midrule
\textbf{\multirow{6}{*}{\rotatebox[origin=c]{90}{STS14}}} & deft-forum & \textbf{44.33} & 39.03 & [0.24, 10.60] & \textbf{52.50} & 46.20 & [2.44, 10.69] & \textbf{45.60} & \textbf{42.95} & [-2.29, 9.03]\\
 & deft-news & \textbf{70.69} & \textbf{68.99} & [-2.55, 5.86] & \textbf{70.64} & \textbf{73.08} & [-5.80, 0.72] & 62.84 & \textbf{67.33} & [-8.61, -0.31]\\
 & headlines & \textbf{65.65} & 61.87 & [1.78, 5.86] & \textbf{68.38} & 66.33 & [0.37, 3.78] & \textbf{62.00} & \textbf{62.09} & [-1.86, 1.88]\\
 & images & \textbf{78.98} & 70.36 & [6.31, 11.03] & \textbf{81.46} & \textbf{80.51} & [-0.52, 2.39] & \textbf{78.33} & \textbf{76.98} & [-0.42, 3.18]\\
 & OnWN & \textbf{69.20} & \textbf{67.45} & [-0.09, 3.68] & 75.92 & \textbf{79.37} & [-4.78, -2.12] & \textbf{75.35} & \textbf{74.69} & [-0.76, 2.17]\\
 & tweet-news & \textbf{74.77} & 71.23 & [0.93, 6.60] & \textbf{76.38} & \textbf{74.89} & [-0.51, 3.83] & \textbf{71.17} & \textbf{68.78} & [-0.32, 5.75]\\
\midrule
\textbf{\multirow{5}{*}{\rotatebox[origin=c]{90}{STS15}}} & answers-forums & \textbf{67.33} & 50.25 & [11.71, 22.87] & \textbf{69.89} & \textbf{68.28} & [-2.58, 5.53] & \textbf{62.27} & 53.74 & [3.31, 13.88]\\
 & answers-students & \textbf{71.28} & \textbf{69.99} & [-1.16, 3.79] & \textbf{73.30} & \textbf{73.95} & [-2.72, 1.46] & \textbf{74.20} & \textbf{72.45} & [-0.12, 3.77]\\
 & belief & \textbf{72.83} & 58.77 & [9.33, 20.04] & \textbf{76.67} & \textbf{73.71} & [-0.03, 6.03] & \textbf{74.17} & 61.73 & [8.05, 18.15]\\
 & headlines & \textbf{72.14} & 69.61 & [0.86, 4.23] & \textbf{74.78} & 72.93 & [0.45, 3.18] & \textbf{68.49} & \textbf{68.58} & [-1.52, 1.40]\\
 & images & \textbf{81.83} & 73.85 & [5.72, 10.37] & \textbf{84.84} & 83.18 & [0.33, 3.02] & \textbf{82.60} & 80.04 & [1.08, 4.07]\\
\midrule
\textbf{\multirow{5}{*}{\rotatebox[origin=c]{90}{STS16}}} & answer-answer & \textbf{61.71} & 43.99 & [10.25, 25.59] & \textbf{66.58} & 54.51 & [6.46, 18.13] & \textbf{58.65} & 43.41 & [8.65, 22.65]\\
 & headlines & \textbf{70.57} & 67.05 & [0.74, 6.55] & \textbf{72.81} & \textbf{71.00} & [-1.11, 4.68] & \textbf{67.87} & \textbf{66.55} & [-1.12, 4.03]\\
 & plagiarism & \textbf{78.02} & 72.25 & [1.47, 10.68] & \textbf{82.97} & \textbf{84.45} & [-4.11, 0.74] & \textbf{80.59} & 75.21 & [1.93, 9.23]\\
 & postediting & \textbf{81.11} & 69.03 & [8.03, 17.65] & \textbf{82.31} & \textbf{82.73} & [-2.80, 2.06] & \textbf{78.97} & 73.87 & [1.53, 9.58]\\
 & question-question & \textbf{66.88} & 58.32 & [1.26, 15.14] & \textbf{74.19} & \textbf{72.29} & [-1.78, 5.20] & \textbf{66.79} & \textbf{63.94} & [-3.32, 9.19]\\
\midrule
\end{tabular}
\captionsetup{width=6in}
\caption{\textbf{MaxPool Spearman vs. MeanPool Spearman:}
		Pearson correlations between human sentence similarity scores and a generated scores.
        Values in bold represent the best result for a subtask given a set of word vectors, based on a 95\% BCa confidence interval~\cite{Efron1987} on the differences between the two correlations.
        In cases of no significant difference, both values are in bold.}

\label{tab:mps-v-asp}
\end{sidewaystable}

\begin{sidewaystable}
	\centering
		\begin{tabular}{s@{\hskip 7pt}k@{\hskip 9pt}ssk@{\hskip 9pt}ssk@{\hskip 9pt}ssk}
			\toprule
			&  & \multicolumn{3}{c}{\textbf{GloVe}} & \multicolumn{3}{c}{\textbf{fastText}} & \multicolumn{3}{c}{\textbf{word2vec}}\\
			\cline{3-11}
			&  & \textbf{MPS} & \textbf{CKA} & \textbf{$\Delta$95\% CI} & \textbf{MPS} & \textbf{CKA} & \textbf{$\Delta$95\% CI} & \textbf{MPS} & \textbf{CKA} & \textbf{$\Delta$95\% CI}\\
			\midrule
			\toprule
			\textbf{\multirow{5}{*}{\rotatebox[origin=c]{90}{STS12}}} & MSRpar & \textbf{40.00} & \textbf{40.01} & [-2.97, 3.13] & \textbf{44.48} & \textbf{44.41} & [-2.53, 2.69] & \textbf{36.70} & \textbf{36.47} & [-2.42, 2.99]\\
			& MSRvid & \textbf{77.66} & \textbf{76.81} & [-0.48, 2.28] & 82.44 & \textbf{84.45} & [-3.06, -1.01] & 74.34 & \textbf{79.95} & [-7.04, -4.33]\\
			& SMTeuroparl & \textbf{46.52} & \textbf{48.94} & [-5.04, 0.24] & \textbf{50.18} & \textbf{51.36} & [-3.20, 0.78] & \textbf{34.13} & \textbf{35.28} & [-3.68, 1.15]\\
			& surprise.OnWN & \textbf{69.23} & \textbf{67.86} & [-0.21, 2.89] & \textbf{73.12} & 70.14 & [1.44, 4.57] & \textbf{69.06} & \textbf{68.19} & [-0.54, 2.37]\\
			& surprise.SMTnews & 49.28 & \textbf{53.80} & [-8.09, -1.03] & \textbf{55.01} & 52.02 & [0.03, 6.21] & 45.09 & \textbf{48.30} & [-6.28, -0.43]\\
			\midrule
			\textbf{\multirow{3}{*}{\rotatebox[origin=c]{90}{STS13}}} & FNWN & \textbf{46.16} & \textbf{36.36} & [-0.46, 20.49] & \textbf{44.14} & \textbf{43.61} & [-8.39, 9.81] & \textbf{49.66} & 40.16 & [0.38, 20.09]\\
			& headlines & \textbf{70.60} & \textbf{71.85} & [-2.83, 0.28] & \textbf{73.04} & \textbf{73.61} & [-2.01, 0.87] & \textbf{65.89} & \textbf{64.66} & [-0.18, 2.77]\\
			& OnWN & \textbf{61.03} & \textbf{60.95} & [-1.90, 2.17] & 71.37 & \textbf{74.25} & [-4.60, -1.38] & 69.40 & \textbf{72.06} & [-4.29, -1.20]\\
			\midrule
			\textbf{\multirow{6}{*}{\rotatebox[origin=c]{90}{STS14}}} & deft-forum & 44.33 & \textbf{50.65} & [-9.84, -2.84] & \textbf{52.50} & \textbf{54.16} & [-4.33, 1.15] & 45.60 & \textbf{52.17} & [-9.61, -3.73]\\
			& deft-news & \textbf{70.69} & \textbf{73.44} & [-5.59, 0.40] & \textbf{70.64} & \textbf{73.06} & [-5.57, 0.50] & 62.84 & \textbf{67.26} & [-7.53, -1.56]\\
			& headlines & \textbf{65.65} & \textbf{66.32} & [-2.24, 0.89] & \textbf{68.38} & \textbf{68.45} & [-1.61, 1.51] & \textbf{62.00} & \textbf{61.54} & [-0.91, 1.93]\\
			& images & \textbf{78.98} & 77.47 & [0.08, 2.95] & \textbf{81.46} & \textbf{81.76} & [-1.42, 0.76] & 78.33 & \textbf{80.57} & [-3.49, -1.07]\\
			& OnWN & \textbf{69.20} & \textbf{69.16} & [-1.44, 1.52] & 75.92 & \textbf{78.46} & [-3.80, -1.36] & 75.35 & \textbf{77.00} & [-2.72, -0.61]\\
			& tweet-news & \textbf{74.77} & \textbf{73.95} & [-0.92, 2.84] & \textbf{76.38} & 73.41 & [1.26, 4.81] & \textbf{71.17} & \textbf{71.86} & [-2.28, 1.03]\\
			\midrule
			\textbf{\multirow{5}{*}{\rotatebox[origin=c]{90}{STS15}}} & answers-forums & \textbf{67.33} & \textbf{66.48} & [-2.52, 4.20] & 69.89 & \textbf{72.78} & [-5.92, -0.14] & \textbf{62.27} & \textbf{64.01} & [-5.00, 1.36]\\
			& answers-students & \textbf{71.28} & \textbf{72.75} & [-3.22, 0.22] & \textbf{73.30} & \textbf{71.92} & [-0.23, 3.11] & \textbf{74.20} & \textbf{73.59} & [-0.86, 2.25]\\
			& belief & \textbf{72.83} & \textbf{71.56} & [-2.26, 5.44] & \textbf{76.67} & \textbf{76.00} & [-1.28, 2.84] & \textbf{74.17} & \textbf{74.16} & [-2.56, 3.22]\\
			& headlines & 72.14 & \textbf{74.05} & [-3.12, -0.71] & \textbf{74.78} & \textbf{75.58} & [-2.01, 0.35] & \textbf{68.49} & \textbf{69.00} & [-1.66, 0.68]\\
			& images & \textbf{81.83} & \textbf{81.35} & [-0.97, 2.00] & \textbf{84.84} & \textbf{85.40} & [-1.52, 0.38] & 82.60 & \textbf{84.02} & [-2.37, -0.50]\\
			\midrule
			\textbf{\multirow{5}{*}{\rotatebox[origin=c]{90}{STS16}}} & answer-answer & \textbf{61.71} & 56.04 & [0.91, 10.55] & \textbf{66.58} & 61.81 & [1.28, 8.66] & \textbf{58.65} & 51.21 & [2.99, 12.68]\\
			& headlines & \textbf{70.57} & \textbf{70.83} & [-2.46, 1.94] & \textbf{72.81} & \textbf{72.87} & [-2.50, 2.12] & \textbf{67.87} & 65.52 & [0.39, 4.59]\\
			& plagiarism & \textbf{78.02} & \textbf{79.36} & [-3.91, 1.41] & \textbf{82.97} & 79.83 & [0.66, 5.92] & \textbf{80.59} & \textbf{80.66} & [-1.99, 1.86]\\
			& postediting & \textbf{81.11} & \textbf{79.94} & [-1.25, 4.32] & \textbf{82.31} & 80.40 & [0.16, 3.85] & \textbf{78.97} & \textbf{78.86} & [-2.01, 2.14]\\
			& question-question & 66.88 & \textbf{72.01} & [-9.29, -1.58] & \textbf{74.19} & \textbf{73.68} & [-2.15, 3.23] & 66.79 & \textbf{70.59} & [-8.03, -0.16]\\
			\midrule
		\end{tabular}
	\captionsetup{width=6in}
	\caption{\textbf{MaxPool Spearman vs. CKA Gaussian:}
	Pearson correlations between human sentence similarity scores and a generated scores. Generated scores
	Values in bold represent the best result for a subtask given a set of word vectors,
	based on a 95\% BCa confidence interval~\cite{Efron1987} on the differences between the two correlations.
	In cases of no significant difference, both values are in bold.}
	\label{tab:mps-v-cka}
\end{sidewaystable}

\end{document}